\documentclass[10pt,twocolumn,letterpaper]{article}
\usepackage[pagenumbers]{cvpr} % To force page numbers, e.g. for an arXiv version

\usepackage[accsupp]{axessibility} % Improves PDF readability for those with disabilities
\usepackage{graphicx}
\usepackage{amsmath}
\usepackage{amssymb}
\usepackage{booktabs, subcaption}
\usepackage{pifont}% http://ctan.org/pkg/pifont
\newcommand{\cmark}{\ding{51}}%
\newcommand{\xmark}{\ding{55}}%

\usepackage[pagebackref,breaklinks,colorlinks]{hyperref}

\usepackage[capitalize]{cleveref}
\crefname{section}{Sec.}{Secs.}
\Crefname{section}{Section}{Sections}
\Crefname{table}{Table}{Tables}
\crefname{table}{Tab.}{Tabs.}

\def\cvprPaperID{9521} % *** Enter the CVPR Paper ID here
\def\confName{CVPR}
\def\confYear{2022}

\usepackage{overpic}
\usepackage{enumitem} %< control spacing in itemize/enumerate/...
\usepackage{overpic} %< add raw math symbols to figures
\usepackage{color}
\definecolor{turquoise}{cmyk}{0.65,0,0.1,0.3}
\definecolor{purple}{rgb}{0.65,0,0.65}
\definecolor{dark_green}{rgb}{0, 0.5, 0}
\definecolor{orange}{rgb}{0.8, 0.6, 0.2}
\definecolor{red}{rgb}{0.8, 0.2, 0.2}
\definecolor{darkred}{rgb}{0.6, 0.1, 0.05}
\definecolor{blueish}{rgb}{0.0, 0.3, .6}
\definecolor{light_gray}{rgb}{0.7, 0.7, .7}
\definecolor{pink}{rgb}{1, 0, 1}
\definecolor{greyblue}{rgb}{0.25, 0.25, 1}

\newcommand{\Fig}[1]{Fig.~\ref{fig:#1}}

\newcommand{\Table}[1]{Table~\ref{tab:#1}}

\newcommand{\Eq}[1]{Eq.~\ref{eq:#1}}

\usepackage{blindtext}

\renewcommand{\paragraph}[1]{\vspace{1em}\noindent\textbf{#1}.}
\begin{document}
\title{CAD: Co-Adapting Discriminative Features for Improved Few-Shot Classification}

\author{Philip Chikontwe, Soopil Kim, Sang Hyun Park \\
Department of Robotics and Mechatronics, DGIST, South Korea\\
{\tt\small \{philipchicco,soopilkim,shpark13135\}@dgist.ac.kr}
% \and
}
\maketitle
\begin{abstract}
Few-shot classification is a challenging problem that aims to learn a model that can adapt to unseen classes given a few labeled samples. Recent approaches pre-train a feature extractor, and then fine-tune for episodic meta-learning. Other methods leverage spatial features to learn pixel-level correspondence while jointly training a classifier. However, results using such approaches show marginal improvements. In this paper, inspired by the transformer style self-attention mechanism, we propose a strategy to cross-attend and re-weight discriminative features for few-shot classification. Given a base representation of support and query images after global pooling, we introduce a single shared module that projects features and cross-attends in two aspects: (i) \textit{query} to \textit{support}, and (ii) \textit{support} to \textit{query}. The module computes attention scores between features to produce an attention pooled representation of features in the same class that is later added to the original representation followed by a projection head. This effectively re-weights features in both aspects (i \& ii) to produce features that better facilitate improved metric-based meta-learning. Extensive experiments on public benchmarks show our approach outperforms state-of-the-art methods by 3\%$\sim$5\%.        
\end{abstract}
\section{Introduction}
\label{sec:intro}
% 1
Deep learning has achieved remarkable success in numerous computer vision tasks, reaching human-level performance in domains with sufficiently large-scale labeled training data. However, large-scale data is not always readily available due to costly curation and labeling. Hence, there is a ongoing research effort to design models that can learn to solve tasks using a limited number of labeled examples to alleviate this requirement. While several approaches such as transfer-, semi- and unsupervised learning have shown reasonable performance, learning to generalize with an extremely limited number of labeled samples is still a challenge. Motivated by the ability of humans to rapidly adapt to new tasks using prior knowledge, the field of few-shot learning shows promise in achieving this goal. 
% 2
\begin{figure}[t]
\begin{center}
\includegraphics[width=\linewidth]{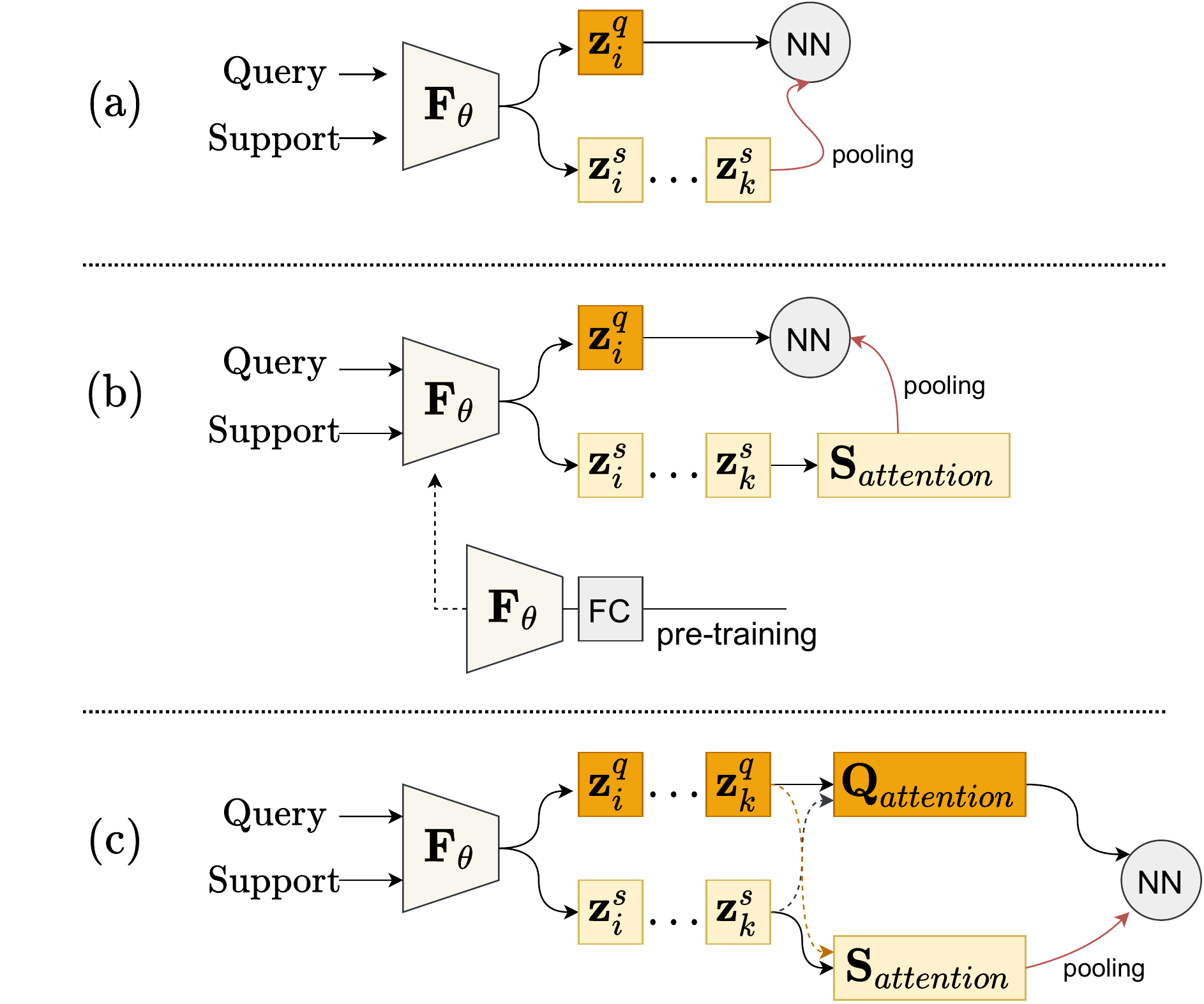}
\end{center}
\caption{
\textbf{Visual intuition} of the proposed framework compared to other approaches using attention. \textbf{(a)} Standard ProtoNet\cite{Snell2017PrototypicalNF} architecture with nearest neighbour classifier (NN). \textbf{(b)} Adaptation of support embeddings only without using relations in the query sets, with backbone pre-training \cite{ye2020few}. \textbf{(c)} Our method leverages attention from each set in a cross-relational perspective to improve classification performance by re-weighting features.
}
\label{fig:teaser}
\end{figure}

Few-shot learning~\cite{koch2015siamese, Ravi2017OptimizationAA, Snell2017PrototypicalNF} aims to classify unlabeled query samples from unseen classes using only limited labeled support examples. Recent approaches have leveraged the meta-learning paradigm where transferable knowledge across a collection of tasks is learned and propagated to improve generalization. In particular, one of the promising approaches have used the meta-learning framework to optimize model parameters~\cite{Finn2017ModelAgnosticMF, Ravi2017OptimizationAA, nichol2018first} with a few gradient steps, thus enabling neural network classifiers to quickly adapt to unseen classes. Other works employ similarity information between images (\Fig{teaser}(a)), augmentation of training data with generative techniques~\cite{Hariharan2017LowShotVR,Schwartz2018DeltaencoderAE}, or two-stage approaches~\cite{chen2019closer, raghu2019rapid, yang2020free, shen2021partial} that involve first pre-training the model on existing known classes and later use a meta-finetuning strategy. In the general meta-setup, tasks defined in the training phase mimic the testing phase to encourage generalization of models. 
% 3

Though pre-training strategies show promising results, Chen \etal\cite{chen2019closer} argued that fine-tuning shows marginal improvements. Moreover, the standard technique of obtaining a base representation with global-average pooling is considered limited, since it is sensitive to object pose and discards key semantic details, and may be difficult to learn generalizable embeddings without being distracted by spurious features~\cite{kang2021relational,hong2021reinforced}. Thus, recent works focus on learning finer distinctions using spatial features only via spatial attention mechanisms~\cite{Hou2019CrossAN} or other forms of relational learning~\cite{wertheimer2021few,kang2021relational} between query and support features to enhance performance. Nevertheless, methods leveraging this setting \ie, spatial feature-based learning, still report marginal improvements over purely discriminative approaches with the exception of some works that provide insight into the model reasoning procedure by visualizing object attended regions for interpretability.
% 4

Here, we argue that meta-learning with instance embeddings only is still viable and can improve few-shot performance when attending to the relevant features. For example, Ye \etal\cite{ye2020few} attempt to address this and show that optimally discriminative features can be meta-learned using a set-to-set function based on the transformer model~\cite{vaswani2017attention} (\Fig{teaser}(b)). The self-attention mechanism in the Transformer presents several desirable properties for set based problems \ie, permutation invariance, interpolation, and contextualization. As a consequence, any meta-learner can leverage such a set-based transformation to adapt instance embeddings relative to other classes for improved prototype generation, or to produce better embeddings. However, while Ye \etal\cite{ye2020few} showed improvements, embedding adaptation is performed on support samples only without leveraging query information, a key distinction with recent spatial techniques using cross-attention~\cite{wertheimer2021few,Hou2019CrossAN,kang2021relational} to learn better correspondences. Moreover, the authors use a transfer-based approach requiring pre-training on base classes before meta-training.      

In this work, we propose an end-to-end meta-learning strategy trained from scratch to co-attend to both support and query features using self-attention (\Fig{teaser}(c)). This is useful to focus on common object features and avoid distractors for better matching. In particular, we introduce a single shared attention module that first projects global pooled base representations and computes attention scores per $k$-shot task via scaled-dot product. Given the scores per shot, we take the mean of the scores to obtain an attention pooled feature (dot product of features and scores) that is later added to the initial projected features before producing the final re-weighted features. Concretely, to improve support features \ie, \textit{query} to \textit{support}, attention scores between query and support features are employed to pool the initial support features producing a support prototype concatenated with the initial features, and vice versa for \textit{support} to \textit{query} to produce a query prototype that improves the initial query features. This strategy is conceptually equivalent to spatial based cross-attention that implicitly re-weights the base spatial maps to attend to relevant object regions. Finally, we employ a nearest neighbor classifier on the refined features for few-shot classification. The contributions of this paper can be summarized as follows:

\begin{itemize}[leftmargin=*]
\setlength\itemsep{-.3em}
\item To improve model-based embedding adaptation, we propose to cross-attend support and query embeddings by re-weighting each instance relative to the other via self-attention.
\item We reveal that implicitly re-weighting features with their prototypes (support/query) via self-attention improves metric based few-shot performance and adds minimal over-head in learnable parameters.
\item Extensive evaluation of the proposed components verify the effectiveness of our method. Moreover, we show competitive results over state-of-the-art methods on several benchmark datasets.
\end{itemize}

\begin{figure*} %[ht!]
\begin{center}
\includegraphics[width=0.9\linewidth, height=7cm]{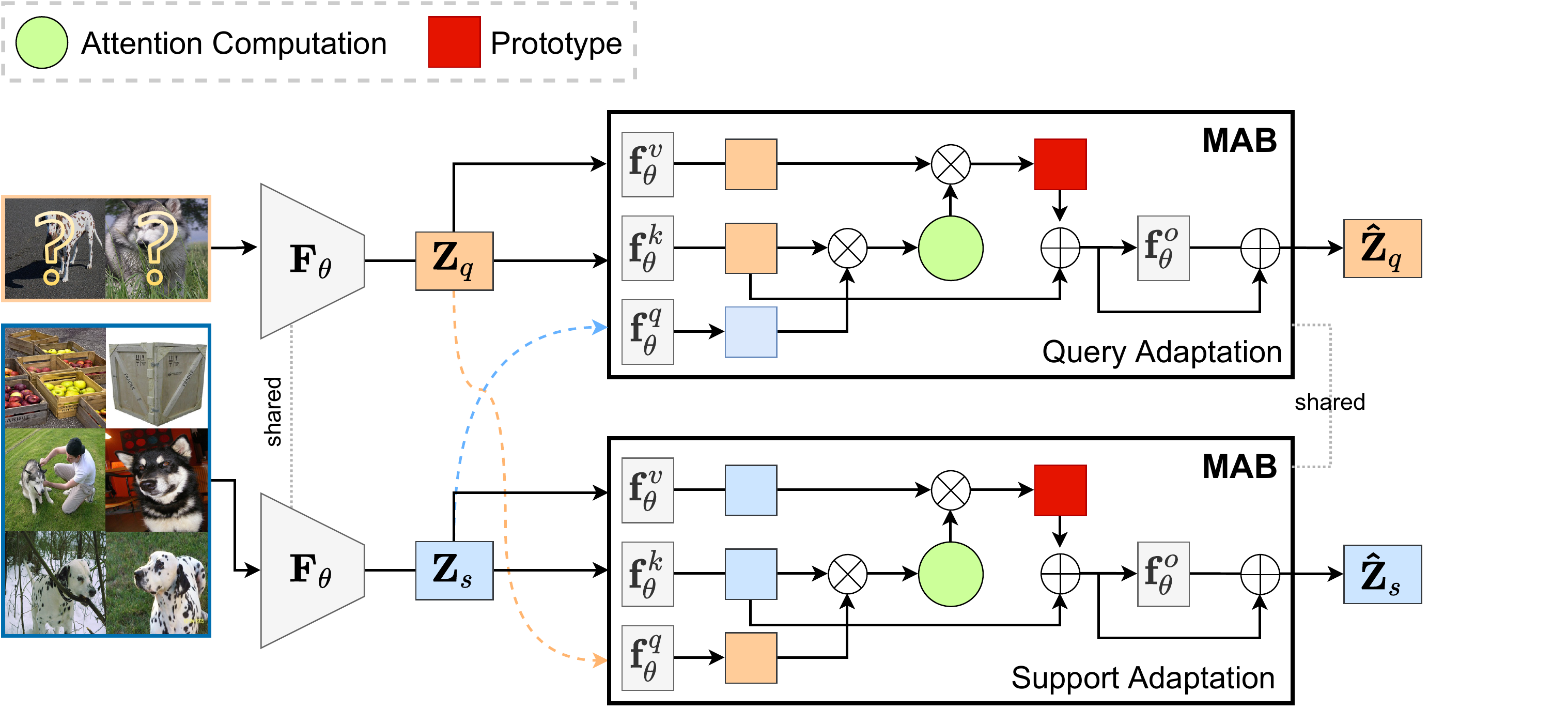}
\end{center}
\caption{
\textbf{Overview of the proposed approach}. We employ episodic meta-training from scratch to learn discriminative co-adapted embedding for support and query images. Concretely, a shared backbone extracts globally pooled features $\mathbf{Z}$ and performs feature adaptation using a shared self-attention module by considering support embeddings $\mathbf{Z_s}$ as keys and value pairs along with query $\mathbf{Z_q}$ to compute attention scores that re-weight $\mathbf{Z_s}$ to produce a prototype $\mathbf{p}$ added to $\mathbf{Z_s}$. Finally, we obtain $\mathbf{\hat{Z}_s}$ after a projection via $\mathbf{f}^{o}_{\theta}$ (the same intuition is used for $\mathbf{Z_q}$ key/value pairs with $\mathbf{Z_s}$ as the query).
}
\vspace{-2mm}
\label{fig:overview}
\end{figure*} 

\section{Related works}
\label{sec:related}

\paragraph{Few-Shot Classification} Several works have been proposed to address few-shot learning~\cite{vinyals2016matching,sung2018learning,Finn2017ModelAgnosticMF,Snell2017PrototypicalNF,ye2020few,liu2020universal,chen2021self,zhang2019variational,li2019few,bateni2020improved,hong2021reinforced}. Existing methods broadly fall into the following categories: model initialization based methods~\cite{Finn2017ModelAgnosticMF,finn2018probabilistic}, metric learning methods~\cite{vinyals2016matching,sung2018learning,Snell2017PrototypicalNF,triantafillou2019meta,bateni2020improved}, and hallucination based methods~\cite{gidaris2018dynamic,luo2021few, li2020adversarial}. The first aims to learn a good model initialization by learning to finetune, where classifiers for unseen classes can be quickly adapted with a small number of gradient steps. However, these approaches have been reported to fail when large domain shifts exist between base and unseen classes~\cite{chen2019closer}. 

On the other hand, metric learning methods address few-shot classification by learning to compare inputs by determining similarity during training. Here, predictions are often conditioned on distance metrics that include cosine similarity~\cite{vinyals2016matching}, euclidean distance to mean class representation~\cite{Snell2017PrototypicalNF}, and relation modules~\cite{sung2018learning}. Along this line of thought, recent approaches adopt a transfer-learning approach \cite{chen2019closer,dhillon2019baseline,gidaris2018dynamic,mangla2020charting,liu2020negative,liu2018learning,rodriguez2020embedding} where pre-training is employed with subsequent fine-tuning as a strong baseline for few-shot learning. Among these, our work falls under metric based approaches, and aims to improve transferability of features by co-adapting with a discriminative cross-attention module.   

\paragraph{Few-shot Learning with Attention} Generally, attention can be employed to reveal the structural layout considering local patch neighborhoods~\cite{shechtman2007matching} of an image, or highlight relevant features in set-based tasks by measuring similarities among inputs~\cite{vaswani2017attention,lee2019set}. This technique has been extensively explored in literature for both 2D and 3D visual tasks~\cite{wang2018non,zoran2020towards,qiu2021geometric}. For example, CrossTransformer~\cite{Doersch2020CrossTransformersSF}, CAN~\cite{Hou2019CrossAN}, and RENet~\cite{kang2021relational} employ attention modules to project query and support features into the space of the other using spatial information. However, these methods incur several parameters over the backbone network and include additional objectives to regularize training.  

Other settings such as incremental few-shot learning also use attention to regularize unseen class features by attending to seen classes~\cite{Ren2019IncrementalFL}. FRN \cite{wertheimer2021few} formulate few-shot classification as a reconstruction problem following the work of Zhang \etal\cite{zhang2020deepemd}. In the context of multi-domain few-shot learning, Liu \etal\cite{liu2020universal} use self-attention to select appropriate representations across domain specific backbones. Also, the benefit of using the Transformer \cite{vaswani2017attention} model is further explored in FEAT \cite{ye2020few}, where self-attention is used as a set-to-set transformation to adapt support representations. Inspired by prior works, we focus on the single domain setting, and propose a hybrid set-to-prototype cross-attention strategy that re-weights support set features by all query embeddings (or vice versa) to produce co-adapted features. Concretely, our approach is conceptually inductive, but can also be considered transductive given the use of unlabelled query samples similar to prior works \cite{bateni2022enhancing,ziko2021transductive}.

\section{Method}
\label{sec:method}

In this section, we introduce the general problem setting for few-shot classification and its related formulations. Following, we describe the overall method as presented in \Fig{overview}, explain the attention mechanism and it's components, including the learning objective.    

\subsection{Problem Setting}
\label{method:setting}
\paragraph{Overview} We consider the standard few-shot learning problem in which we are given a labeled train set $\mathcal{D}_{t}$, an unlabeled query set $\mathcal{D}_{q}$, and a few labeled examples from a support set $\mathcal{S}$ sharing the same label space with the query set. The label space of the training and query set are non-overlapping \ie, $\{Y_t \cap Y_q\} = \emptyset$, where $\{Y_t\}$ and $\{Y_q\}$ denote the training and query labels, respectively. Thus, given a set of labeled tasks from $\mathcal{D}_{t}$, and a few examples from the support set; the goal is to train a model that can generalize to novel tasks in the query set $\mathcal{D}_{q}$. 

Towards this goal, meta-learning with episodic training can be employed. This scheme aims to improve generalization by mimicking the low-data setting encountered in inference by creating balanced episodes. An episode is formed by sampling disjoint data points from the support set consisting a few labeled points, and the query set where labels are used to calculate prediction errors per episode. Moreover, each episode defines an "$N$-way,$k$-shot" task, where $N$ indicates the number of classes per episode, and $k$ is the number of support examples per class. 

\paragraph{Prototypical Baselines} In the context of Prototypical networks~\cite{Snell2017PrototypicalNF}, one aims to construct an embedding space in which points cluster around a single \textit{prototype} $\mathbf{p}_k \in \mathbb{R}^Z$  representation of each class $k$. Here, the objective is to learn an embedding function $\mathbf{F}_{\theta}(\text{x}): \text{x} \in \mathbb{R}^{H \times W \times C} \rightarrow \mathbb{R}^Z$ to transform an input into a $\mathbf{z}$-dimensional vector. To convey the description of the class as meta-data, the prototype of each class $\mathbf{p}_k$ is defined as the average of all embedded samples in that class:

\begin{equation}
	\mathbf{p}_k = \frac{1}{|S_k|} \sum_{(\text{x}_i,y_i) \sim S_k}^{} \mathbf{F}_{\theta}(\text{x}_i), \label{eq:1}
\end{equation} 

\noindent where $S_k$ denotes samples from class $k$. To obtain the probability distribution over classes for a query sample $\text{x}_q$, the softmax function on distances of the query to the prototypes is employed:
\begin{equation}
	p_{\theta}(y_i = k|\text{x}_q) = \frac{\text{exp}(-d(\mathbf{F}_{\theta}(\text{x}_q),\mathbf{p}_k))}{\sum_{k^\prime}^{} \text{exp}(-d(\mathbf{F}_{\theta}(\text{x}_q),\mathbf{p}_{k^\prime}))}, \label{eq:2}
\end{equation}

\noindent where $d(\cdot{})$ is a distance function between query and the prototype. Consequently,  $p_{\theta}(y_i = k|\text{x}_q)$ is used to assign the correct class per sampled task, and the model learns to minimize the predictive error on query samples per task. 

Though effective in practice, we argue that the prototype embeddings are not ideal to learn a discriminative model. Since every element in $S_k$ is processed independently in the pooling operation (\ie via mean), some information regarding interactions between them may be discarded making it difficult for the model to learn robust embeddings.  In what follows, we describe how to co-adapt the embeddings via a transformer style set-to-prototype function using self-attention.  

\subsection{Instance Embedding Adaptation}
\label{method:coadapt}

In order to co-adapt support and query embeddings $\text{x}_s$ and $\text{x}_q$, we introduce an additional step where the initial embeddings are further transformed using self-attention. Here, attention is employed to find the relevant features to attend to in $\text{x}_s$ and $\text{x}_q$ by considering discriminative feature similarity only, and thus enables us to re-weight the features. Formally, let $z_q = \mathbf{F}_{\theta}(\text{x}_q)$ and $z_s = \mathbf{F}_{\theta}(\text{x}_s)$ be the initial query and support embeddings after the penultimate layer, each having vector dimension $m$ \ie, $z \in \mathbb{R}^m $. Our goal is to obtain adapted embeddings $z^{\prime}_q$ and $z^{\prime}_s$. 

\paragraph{Self-Attention} Following the work of Vaswani \etal\cite{vaswani2017attention} and its related extensions presented by Lee \etal\cite{lee2019set} for set-based tasks, we employ an attention function $\varphi(\mathbf{Q},\mathbf{K},\mathbf{V})$ that measures similarity between a query vector $\mathbf{Q}$ with key-value pairs $\mathbf{K} \in \mathbb{R}^{n \times m}$ and $\mathbf{V} \in \mathbb{R}^{n\times m}$ via: 

\begin{equation}
	\varphi(\mathbf{Q},\mathbf{K},\mathbf{V}) = \text{softmax}(\frac{\mathbf{Q}\mathbf{K}^T}{\sqrt{m}})\mathbf{V}. \label{eq:3}
\end{equation}

Concretely, the formulation introduced by Vaswani \etal\cite{vaswani2017attention} was extended to a multi-head attention block (MAB) where vectors $\mathbf{Q},\mathbf{K},\mathbf{V}$ are first projected onto $h$ different dimensional vectors. Here, $\varphi(\mathbf{Q},\mathbf{K},\mathbf{V})$ then becomes $\varphi(\mathbf{Q}W_j^{\mathbf{Q}},\mathbf{K}W_j^{\mathbf{K}},\mathbf{V}W_j^{\mathbf{V}})$, with each transformation having its own learnable parameters $W$. In this setting, each $W$ can be modeled by functions $\mathbf{f}^{q}_{\theta}$, $\mathbf{f}^{k}_{\theta}$ and $\mathbf{f}^{v}_{\theta}$, respectively. Also, an additional function $\mathbf{f}^{o}_{\theta}$ takes the output of $\varphi$ and adds the residual followed by a layer normalization operation. Mathematically,

\begin{equation}
	\varphi^o = \varphi(\mathbf{Q},\mathbf{K},\mathbf{V}) + \mathbf{Q}, \label{eq:4}
\end{equation}

\begin{equation}
	\text{MAB}_h(\mathbf{Q},\mathbf{K},\mathbf{V}) = \Phi(\varphi^o + \gamma(\mathbf{f}^{o}_{\theta}(\varphi^o))) \label{eq:5}
\end{equation}

\noindent where $\gamma$ is the ReLU function and $\varphi^o$ is the output of applying attention on query key-value pairs after individual feature projection using functions $\mathbf{f}^{q,k,v}_{\theta}$ along with the concatenation of $\mathbf{Q}$. $\text{MAB}_h$ defines a single attention block ($h=1$) with an optional normalization layer $\Phi$. It is worth noting that the $\mathbf{Q}$ notation used here is specifically related to features in the $\text{MAB}_h$ module, and should be considered separate from the notion of queries in the few-shot setting.

\begin{figure}[h]
	\begin{center}
		\includegraphics[width=\linewidth]{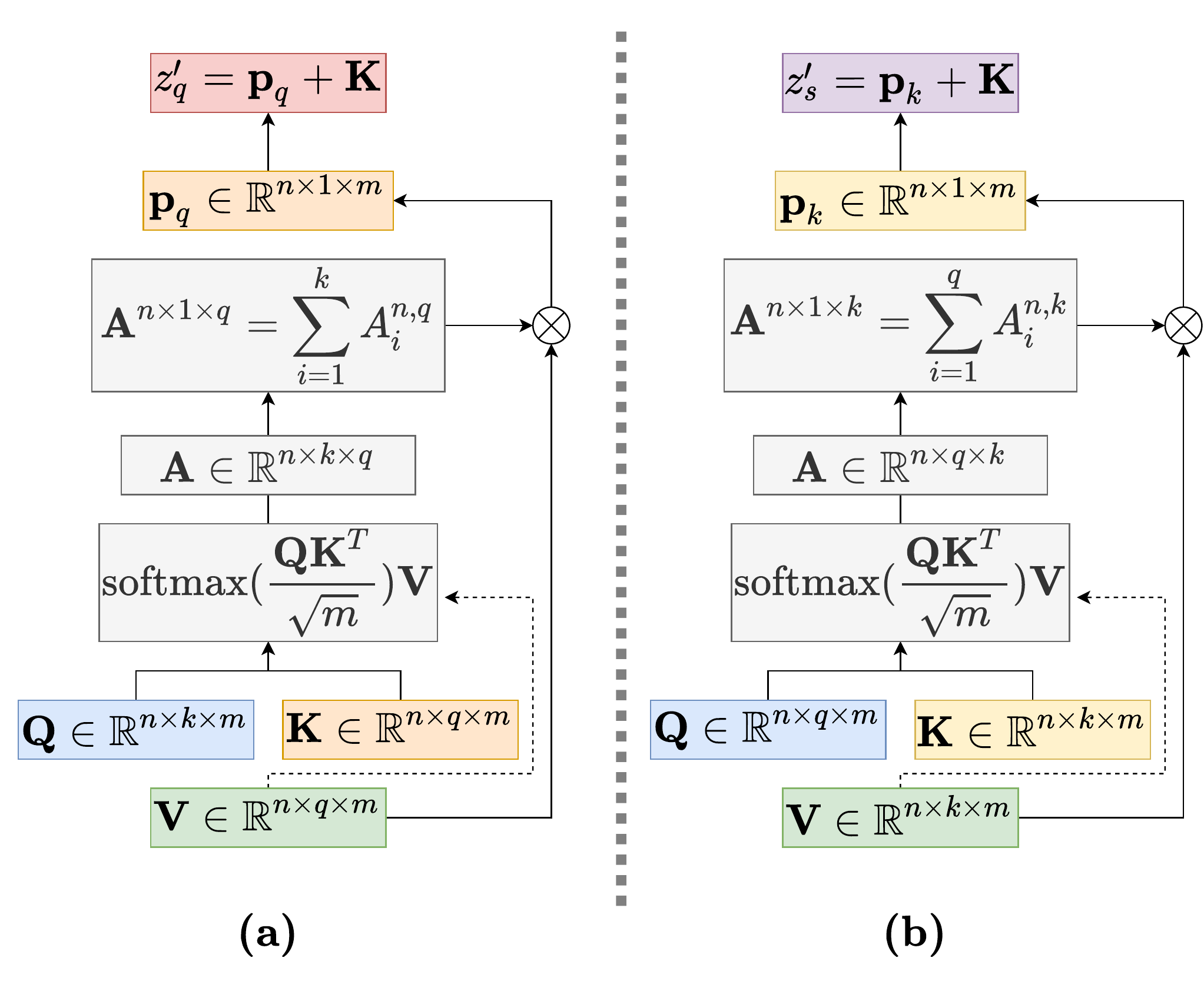}
	\end{center}
	\vspace{-3mm}
	\caption{
		\textbf{Illustration of embedding adaptation}. \textbf{(a)} For query features, $\mathbf{Q} = z_s$ and $\mathbf{K} = \mathbf{V}  = z_q$, where $\mathbf{p}_{k}$ and $\mathbf{p}_{q}$ are the support and query prototypes obtained via dot product of the average attention scores per-$k$ shot with $\mathbf{V}$. \textbf{(b)}: Support features are modified by setting $\mathbf{Q} = z_q$ and $\mathbf{K} = \mathbf{V}  = z_s$. Finally, $n$ denotes the number of classes, with $q$ the number of queries.
	}
	\label{fig:selfatt}
	\vspace{-3mm}
\end{figure}

\paragraph{Cross-Adaptation with MAB} In the context of this work, to adapt features relative to the other, we consider two aspects: (i) \textit{support}-\textit{query}, and (ii) \textit{query}-\textit{support}, as presented in \Fig{selfatt}. For scenario (i), we employ the attention module as $\text{MAB}_h(z_s,z_q,z_q)$ (\Fig{selfatt}(a)), where support features $z_s$ are considered as the query ($\mathbf{Q}$), and the initial query embeddings $z_q$ are key-value pairs. For (ii), the opposite is applied, \ie $\text{MAB}_h(z_q,z_s,z_s)$ (\Fig{selfatt}(b)). Intuitively, this should produce embeddings that are weighted by attention reflecting feature similarity to obtain $z^{\prime}_s = \text{MAB}_h(z_q,z_s,z_s)$, and $z^{\prime}_q = \text{MAB}_h(z_s,z_q,z_q)$, respectively.

To this end, we wish to note that Lee \etal\cite{lee2019set} employ \Eq{4} \& \Eq{5} by making $\mathbf{Q}$ a learnable parameter in order to perform attention-based pooling in set-based tasks \ie $(\mathbf{K} \in \mathbb{R}^{n\times k\times m},\mathbf{V}\in \mathbb{R}^{n\times k\times m})$ are a set of features and $\mathbf{Q} = W \in \mathbb{R}^{n\times m}$ is a tensor initialized from a normal distribution to produce a pooled tensor $\mathbf{O}\in \mathbb{R}^{n\times m}$, with $n$ as the batch dimension, and $k$ the number of samples in the set, respectively. This approach has been shown to produce better pooled embeddings over simply employing the mean directly. In this work, employing \Eq{4} reveals a tensor shape mismatch since few-shot inputs consist $q$-queries and $k$-shot samples for the support set in a meta-setup. To address this, we modify \Eq{4} and \Eq{5} as: 

\begin{equation}
	\hat{\varphi}^o = \varphi(\mathbf{Q},\mathbf{K},\mathbf{V}) + \mathbf{K}, \label{eq:6}
\end{equation}

\begin{equation}
	\text{MAB}_h(\mathbf{Q},\mathbf{K},\mathbf{V}) = \Phi(\hat{\varphi}^o + \gamma(\mathbf{f}^{o}_{\theta}(\hat{\varphi}^o))). \label{eq:7}
\end{equation}

In addition, note that $\varphi(\cdot)$ will result in different attention scores for \textit{query}-\textit{support} and \textit{support}-\textit{query} \ie, $\mathbf{A}_s \in \mathbb{R}^{n\times q\times k}$ and $\mathbf{A}_q \in \mathbb{R}^{n\times k\times q}$ for $k$-shot with $q$ query examples. Thus, in each setting we take the average over $k$ or $q$ to obtain the final scores. For example, in support adaptation, this implies we obtain $\mathbf{A}_s \in \mathbb{R}^{n\times 1\times k}$ and taking the dot product with $\mathbf{V}$ produces a prototype embedding, vice versa for queries. The support/query prototypes reflect the relevant features later added to the initial $\mathbf{K}$ to obtain the final embeddings. Consequently, the prototype of the adapted support embeddings are now obtained via: 

\begin{equation}
	 \mathbf{\hat{p}}_k = \frac{1}{|S_k|} \sum_{(\text{x}_i,y_i) \sim S_k}^{} z^{\prime}_s,
	 \label{eq:8}
\end{equation} 

\noindent and the distribution over classes is:

\begin{equation}
	p_{\theta}(y_i = k|\text{x}_q) = \frac{\text{exp}(-d(z^{\prime}_q,\mathbf{\hat{p}}_k))}{\sum_{k^\prime}^{} \text{exp}(-d(z^{\prime}_q,\mathbf{\hat{p}}_{k^\prime}))}. \label{eq:9}
\end{equation} 

To optimize the backbone and proposed modules, we train the model following the standard setting proposed in Prototypical networks (inductive only) using a cross-entropy loss on the negative distances to minimize the query prediction error per task. During inference, all modules are used as is, maintaining the design of the training phase without omission of any components.

\begin{table*}[t!]
\centering
%\resizebox{\linewidth}{!}{ %< auto-adjusts font size to fill line
\scalebox{0.82}{
\begin{tabular}{@{}lc@{} c c c c c @{}}
\toprule

&&& \multicolumn{2}{c}{\bf{\textit{mini}ImageNet}} & \multicolumn{2}{c}{\bf{\textit{tiered}ImageNet}} \\
Method & Backbone && $1-$shot & $5-$shot & $1-$shot & $5-$shot \\
\midrule 
ProtoNet \cite{Snell2017PrototypicalNF} & ResNet12 && 62.39$\pm$0.21 & 80.53$\pm$0.14 &68.23$\pm$0.23&84.03$\pm$0.16 \\
MetaOptNet \cite{Lee2019MetaLearningWD} & ResNet12 && 62.64$\pm$0.82 & 78.63$\pm$0.46 &65.99$\pm$0.72&81.56$\pm$0.53 \\
SimpleShot \cite{Wang2019SimpleShotRN}  & ResNet18 && 62.85$\pm$0.20 & 80.02$\pm$0.14 &-&- \\
MatchNet \cite{vinyals2016matching}  & ResNet12 && 63.08$\pm$0.80 & 75.99$\pm$0.60 &68.50$\pm$0.92&80.60$\pm$0.71 \\
RFS-simple \cite{tian2020rethinking}  & ResNet12 && 62.02$\pm$0.63 & 79.64$\pm$0.44 &69.74$\pm$0.72&84.41$\pm$0.55 \\
S2M2 \cite{mangla2020charting}  & ResNet34 && 63.74$\pm$0.18 & 79.45$\pm$0.12 &-&- \\
NegMargin \cite{liu2020negative}  & ResNet12 && 63.85$\pm$0.81 & 81.57$\pm$0.56 &-&- \\
CTM \cite{li2019finding}  & ResNet18 && 64.12$\pm$0.82 & 80.51$\pm$0.13 &68.41$\pm$0.39&84.28$\pm$1.73 \\
CAN \cite{Hou2019CrossAN}  & ResNet12 && 63.85$\pm$0.48 & 79.44$\pm$0.34 &69.89$\pm$0.51&84.23$\pm$0.37 \\
DeepEMD \cite{zhang2020deepemd}  & ResNet12 && 65.91$\pm$0.82 & 82.41$\pm$0.56  &71.16$\pm$0.87&86.03$\pm$0.58 \\
FEAT \cite{ye2020few}  & ResNet12 && 66.78$\pm$0.20 & 82.05$\pm$0.14  &70.80$\pm$0.23&84.79$\pm$0.16\\
RENet \cite{kang2021relational}  & ResNet12 && 67.60$\pm$0.44 & 82.58$\pm$0.30  &71.61$\pm$0.51&85.28$\pm$0.35 \\
FRN \cite{wertheimer2021few} & ResNet12 && 66.45$\pm$0.19 & 82.83$\pm$0.13  &72.06$\pm$0.22&86.89$\pm$0.14 \\ 

EPNet \cite{rodriguez2020embedding} & ResNet12 && 66.50$\pm$0.89 & 81.06$\pm$0.61  &76.53$\pm$0.87&87.32$\pm$0.64 \\
Dist-Calib \cite{yang2020free} & WRN28 && 68.51$\pm$0.55 & 82.88$\pm$0.42  &78.19$\pm$0.25&89.90$\pm$0.41 \\
SLK-MS \cite{ziko2021transductive} & ResNet18 && 73.10 & 82.82 &\bf79.99 &86.55 \\
EPNet \cite{rodriguez2020embedding} & WRN28 && 70.74$\pm$0.85 & 84.34$\pm$0.53  &78.50$\pm$0.91&88.36$\pm$0.57 \\
SLK-MS \cite{ziko2021transductive} & WRN28 && 75.17 & 84.28  &81.13 & 87.69 \\
\midrule
Ours   & ResNet12 && \bf77.56$\pm$0.72 & \bf87.68$\pm$0.57 &77.55$\pm$0.74& \bf90.73$\pm$0.54 \\
\bottomrule
\end{tabular}
} % \resizebox
\caption{
Few-shot classification accuracy on \textit{mini}ImageNet and \textit{tiered}ImageNet in the $5$-way $k$-shot setting (mean accuracy in a $\pm95\%$ confidence interval is reported)
} % \caption
\label{tab:1}
\vspace{-2mm}
\end{table*}

\section{Experiments}
\label{sec:experiments}

In this section, details on the datasets employed along with implementation settings are presented. We evaluate our proposed method on standard few-shot benchmarks and compare results with recent state-of-the-art methods. 

\paragraph{Datasets} To verify the proposed method, we address few shot classification under two scenarios: generic object recognition and fine-grained image classification. For this purpose, we employ the \textit{mini}ImageNet~\cite{Ravi2017OptimizationAA,deng2009imagenet}, \textit{tiered}ImageNet~\cite{ren2018meta}, CIFAR-FS~\cite{bertinetto2018meta}, and CUB-200-2011~\cite{wah2011caltech} datasets. 

\textbf{\textit{mini}ImageNet} consists of a subset of 100 object classes from ImageNet~\cite{deng2009imagenet} with 600 images per class. We follow the setting proposed by Ravi \etal\cite{Ravi2017OptimizationAA} to randomly select $64$ base, $16$ validation, and $20$ novel classes, respectively. \textbf{\textit{tiered}ImageNet} is a larger subset of ImageNet with 351/97/160 sub-classes for training/validation/testing stemming from 20/6/8 super-classes. It is a more challenging dataset as the splits are disjoint in terms of super-classes and requires better generalization. \textbf{CUB-200-2011} consists of $200$ classes with a total of $11,788$ images. Following the protocols of Hilliard \etal\cite{hilliard2018few}, the dataset is randomly split into $100$ base, $50$ validation, and $50$ novel classes. \textbf{CIFAR-FS} is built upon CIFAR-100 and has $100$ classes, each containing $600$ images. We use the same split of 64/16/20 as in Bertinetto \etal\cite{bertinetto2018meta}.

\paragraph{Implementation Details} Following recent works~\cite{zhang2020deepemd,kang2021relational,Hou2019CrossAN}, we adopt ResNet12~\cite{he2016deep} as the backbone network. The backbone take images of size $84\times84$ as input and produces embeddings $z_{q,s} \in \mathbb{R}^{640}$ after global pooling. We train our model in the `$5$-way $1$-shot' and `$5$-way $5$-shot' setting with standard normalization and augmentation techniques as in prior work \cite{kang2021relational}. Moreover, our attention block is shared between adaptation steps and only uses a single head (h=1) with dimensions of the fully-connected layers set to $640$, and $d(\cdot{})$ is the euclidean function. In training, Adam optimizer was used with a learning rate of $0.003$ without decay or scheduling. $1$-shot models were trained for $300$ epochs with $200$ tasks sampled per epoch, resulting in $60,000$ tasks. On the other hand, $5$-shot models were trained for $200$ epochs totaling $40,000$ tasks similar to Chen \etal\cite{chen2019closer}. During evaluation, we meta-test 15 query samples per class in each episode and report the average accuracy with 95\% confidence interval over $2,000$ randomly sampled test episodes.

% Subtable
\begin{table}[t!]
	\centering
	\begin{subtable}{.5\textwidth}
	%\centering
	\scalebox{0.90}{
	\begin{tabular}{lccc}
		\toprule
		Method & Backbone & $1-$shot & $5-$shot \\
		\midrule
		Cosine \cite{chen2019closer}  & ResNet34 & 68.00$\pm$0.83 & 84.50$\pm$0.51 \\
		MatchNet \cite{vinyals2016matching}  & ResNet12 & 71.87$\pm$0.85 & 85.08$\pm$0.57 \\
		NegMargin \cite{liu2020negative}  & ResNet18 & 72.66$\pm$0.85 & 89.40$\pm$0.43 \\
		S2M2 \cite{mangla2020charting}  & ResNet18 & 71.81$\pm$0.43 & 86.22$\pm$0.53 \\
		S2M2 \cite{mangla2020charting}  & ResNet34 & 72.92$\pm$0.83 & 86.55$\pm$0.51 \\
		FEAT* \cite{ye2020few}  & ResNet12 & 73.27$\pm$0.22 & 85.77$\pm$0.14 \\
		DeepEMD \cite{zhang2020deepemd}  & ResNet12 & 75.65$\pm$0.83 & 88.69$\pm$0.50 \\
		ProtoNet \cite{Snell2017PrototypicalNF} & ResNet12 & 66.09$\pm$0.92 & 82.50$\pm$0.58 \\
		RENet \cite{kang2021relational}  & ResNet12 & 79.49$\pm$0.44 & 91.11$\pm$0.24 \\
		SLK-MS \cite{ziko2021transductive} & ResNet18 & 81.88 & 88.55 \\
		EPNet \cite{rodriguez2020embedding} & ResNet12 & 82.85$\pm$0.81 & 91.32$\pm$0.41 \\
		FRN \cite{wertheimer2021few} & ResNet12 & \bf83.55$\pm$0.19 & \bf92.92$\pm$0.10 \\
		\midrule
		Ours   & ResNet12 & 82.95$\pm$0.67 & 90.80$\pm$0.51 \\
		\bottomrule
	\end{tabular}
	}
	\caption{\small Results on CUB-200-2011 dataset.}
	\end{subtable}
	
	\centering
	\begin{subtable}{.5\textwidth}
	%\centering
	\scalebox{0.92}{
	\begin{tabular}{lccc}
		\toprule
		Method & Backbone & $1-$shot & $5-$shot \\
		\midrule
		Cosine \cite{chen2019closer} & ResNet34 & 60.39$\pm$0.28 & 72.85$\pm$0.65 \\
		S2M2 \cite{mangla2020charting}  & ResNet18 & 63.66$\pm$0.17&76.07$\pm$0.19 \\
		S2M2 \cite{mangla2020charting}  & ResNet34 &  62.77$\pm$0.23&75.75$\pm$0.13 \\
		ProtoNet \cite{Snell2017PrototypicalNF} & ResNet12 & 72.20$\pm$0.70&83.50$\pm$0.50 \\
		MetaOptNet \cite{Lee2019MetaLearningWD} & ResNet12 &72.80$\pm$0.70& 85.00$\pm$0.50 \\
		Boosting \cite{gidaris2019boosting} & WRN28 & 73.60$\pm$0.30&86.00$\pm$0.20 \\
		RENet \cite{kang2021relational}  & ResNet12 & 74.51$\pm$0.46&86.60$\pm$0.32 \\
		\midrule
		Ours   & ResNet12 & \bf79.97$\pm$0.72 & \bf94.13$\pm$0.41 \\
		\bottomrule
	\end{tabular}
	}
	\caption{\small Results on CIFAR-FS dataset.}
	\end{subtable}
	\caption{
		Few-shot classification accuracy on CUB-200-2011 and CIFAR-FS in the $5$-way $k$-shot setting (mean accuracy in the $\pm95\%$ confidence interval). ``*'' : denotes results reproduced by \cite{kang2021relational}.
	}
	\label{tab:2}
	\vspace{-3mm}
\end{table}

\paragraph{Comparison with State-of-the-Art Methods} Tables \ref{tab:1} and \ref{tab:2} compare the proposed method with current state-of-the-art few-shot methods \cite{kang2021relational,wertheimer2021few,Hou2019CrossAN,zhang2020deepemd,ye2020few}. Our approach shows consistent improvements over several methods in all evaluated settings. Notably, we observe significant gains in both $1$- and $5$-shot settings, highlighting the effectiveness of our approach despite its simplicity. For example, on \textit{mini}ImageNet, our approach shows $+3\%$ and $+4\%$ gain in $1$-shot and $5$-shot settings over the best method. We observed similar results on the challenging \textit{tiered}ImageNet with $+1\%$ gains, except for CUB where our approach shows results on par with current state-of-the-art methods. While closely related method RENet~\cite{kang2021relational} employs spatial relational learning of query/support features, our technique solely benefits from discriminative co-adaptation to enhance performance. Recent work FRN~\cite{wertheimer2021few} equally shows competitve performance, however FRN reformulates few-shot classification to leverage reconstruction of related features, and requires a two-stage process with pre-training followed by episodic fine-tuning. In contrast, our approach can be trained end-to-end from scratch with episodic training.

As shown in Table \ref{tab:2}, our method is highly competitive to the closely related method FEAT \cite{ye2020few}. Note that FEAT is another transfer learning method leveraging self-attention on support samples only, and uses extra regularization terms in the episodic training step. Herein, the results validate our initial hypothesis that discriminative learning alone can still show gains over existing baselines. On the other hand, while our approach does not report the best result on CUB-200, we are on par with the best method FRN and produce the best performance on CIFAR-FS ($\approx 5\%$). In addition, while approaches such as Boosting~\cite{gidaris2019boosting} use self-supervision and auxiliary losses with extra unlabeled samples for semi-supervised few-shot learning, we show that a smaller backbone is still competitive.

\section{Ablation Study}
\label{sec:results}

In this section, we further validate the proposed method by analyzing the effect of omitting certain components/modules on model performance, and also investigate the challenging cross-domain generalization setting. Moreover, we provide quantitative evaluation on the benefit of the proposed attention strategy in a Prototypical baseline. 

\subsection{Improved Prototypical Baseline}
\begin{table}[!h]
	\centering
	\scalebox{0.86}{
		\begin{tabular}{lccc}
			\toprule
			\bf{Model} & Backbone & $1$-shot & $5$-shot \\
			\midrule
			ProtoNet*\cite{Snell2017PrototypicalNF} & ResNet12&51.61$\pm$0.44& 72.28$\pm$0.36 \\
			ProtoNet** w/ \Eq{6} &ResNet12& \bf59.57$\pm$0.61 & \bf84.29$\pm$0.40 \\
			\bottomrule
		\end{tabular}
	}
	\caption{
		Performance comparison of standard ProtoNet baseline and a ProtoNet using self-attention on \textit{mini}ImageNet dataset in the $5$-way $k$-shot scenario. ``*'': denotes reproduced results and ``**'' denotes a model using a non-parametric version of the proposed method following \Eq{6}.
	} 
	\label{tab:base}
\end{table}

\Table{base} summarizes the benefit of employing the proposed cross adaptation of embeddings in ProtoNet\cite{Snell2017PrototypicalNF}. We re-train the ProtoNet model with a non-parametric version of the proposed method \ie, no use of fully-connected layers $\mathbf{f}^{q}_{\theta}$, $\mathbf{f}^{k}_{\theta}$, $\mathbf{f}^{v}_{\theta}$ and $\mathbf{f}^{o}_{\theta}$. In particular, given the base representation after global average pooling, we directly apply \Eq{6} for both support-query and query-support adaptation, without ReLU or normalization layers. Results show that simply re-weighting features is beneficial, especially in the $5$-shot setting where larger gains were observed.

\subsection{Cross-Domain Few-Shot Classification}

\begin{table}[!h]
	\centering
	\scalebox{0.90}{
	\begin{tabular}{lccc}
		\toprule
		\bf{Model} & Backbone & $1$-shot & $5$-shot \\
		\midrule
		ProtoNet* \cite{Snell2017PrototypicalNF} & ResNet18& - & 62.02$\pm$0.70 \\
		SimpleShot* \cite{Wang2019SimpleShotRN} & ResNet18 & 48.56 & 65.63 \\
		MatchNet*  \cite{vinyals2016matching} &ResNet10& 36.61$\pm$0.53 & 55.23$\pm$0.83 \\
		NegMargin* \cite{liu2020negative} &ResNet18 & - & 69.30$\pm$0.70 \\
		MetaOptNet* \cite{Lee2019MetaLearningWD} &ResNet12 & 44.79$\pm$0.75 & 64.98$\pm$0.68 \\
		ASL \cite{afrasiyabi2020associative}   &ResNet18& 46.85$\pm$0.75 & 70.37$\pm$1.02 \\
		FRN \cite{wertheimer2021few}  &ResNet12& 54.11$\pm$0.19 & 77.09$\pm$0.15 \\
		\midrule
		Ours &ResNet12& \bf74.10$\pm$0.75 & \bf86.37$\pm$0.60 \\
		\bottomrule
	\end{tabular}
	}
	\caption{
		Performance comparison in the cross-domain setting: \textit{mini}ImageNet $\rightarrow$ CUB-200-2011 in the $5$-way $k$-shot scenario. ``*'' : denotes results reported by \cite{wertheimer2021few}.
	} 
	\label{tab:3}
\end{table}

In Table \ref{tab:3}, we evaluate on the challenging cross-domain setting of \textit{mini}ImageNet to CUB. Following the splits introduced by Chen \etal\cite{chen2019closer}, we trained our model on \textit{mini}ImageNet only \{base+val+test\}, and meta-test on CUB test split. We report significant improvements over related methods especially in the $1$-shot setting. Our intuition is that in the cross-domain setup, cross adaptation of features leads to better generalization by avoiding distracting features via attention.

\subsection{Effect of the Self-Attending Modules}
% \checkmark
\begin{table}[!h]
	\centering
	\scalebox{0.90}{
	\begin{tabular}{ccccc}
		\toprule
		$\text{MAB}^{S}_h$&$\text{MAB}^{Q}_h$&$\text{MAB}^{*}_h$& $1$-shot & $5$-shot \\
		\midrule
		\cmark&\xmark&\xmark &56.00$\pm$0.79 & 85.10$\pm$0.61 \\
		\xmark&\cmark&\xmark &76.33$\pm$0.73 & 80.10$\pm$0.74 \\
		\xmark&\xmark&\cmark &56.01$\pm$0.86 & 79.94$\pm$0.72 \\
		\midrule
		\cmark&\cmark&\xmark &\bf77.56$\pm$0.72 & \bf87.68$\pm$0.57 \\
		\bottomrule
	\end{tabular}
	}
	\caption{
		Evaluation on the contribution of attention modules in the proposed method with ResNet12 on \textit{mini}ImageNet. $\text{MAB}^{*}_h$ denotes using self-attention only on both query and support features i.e., $z^{\prime}_s = \text{MAB}_h(z_s,z_s,z_s)$ \& $z^{\prime}_q = \text{MAB}_h(z_q,z_q,z_q)$. $\text{MAB}^{S}_h$ implies support adaptation: $z^{\prime}_s = \text{MAB}_h(z_q,z_s,z_s)$, whereas $\text{MAB}^{Q}_h$ is query adaptation  $z^{\prime}_q = \text{MAB}_h(z_s,z_q,z_q)$, respectively.
	}
	\label{tab:effectmodules}
\end{table}
\begin{figure*}
	\centering
	%\scalebox{0.86}{
		\begin{center}
			\includegraphics[width=0.85\linewidth,height=7.0cm]{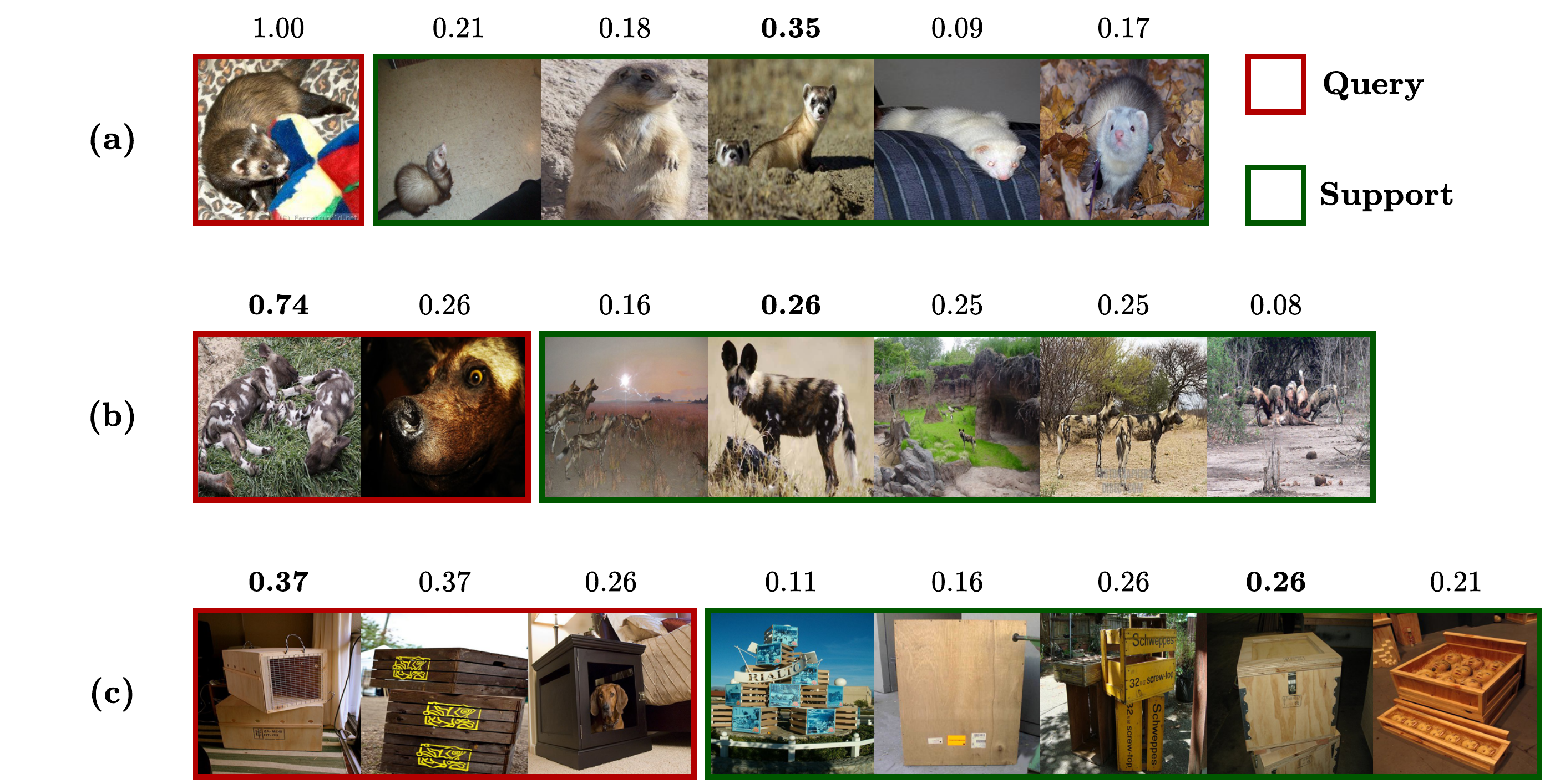}
		\end{center}
	%}
	%\vspace{-5mm}
	\caption{
		\textbf{Illustration of the attention mechanism}. \textbf{(a)} Attention scores of our $5$-way $5$-shot model on the support set using a single query image on a randomly sampled task \textit{mini}ImageNet testing split. \textbf{(b)} Scores for a task with $2$-query images each re-weighted relative to the support samples, and \textbf{(c)} shows scores using $3$-query images.
	}
	\label{fig:qualityatt}
	\vspace{-1mm}
\end{figure*}
\begin{figure}
	\centering
	\setlength{\tabcolsep}{1pt}
	\begin{tabular}{cccc}
		\includegraphics[width=20mm]{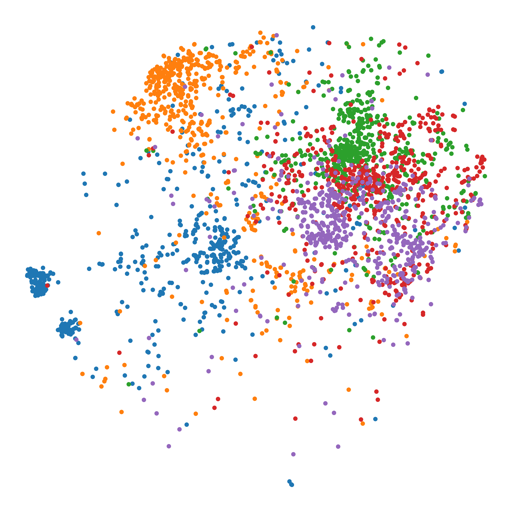} &
		\includegraphics[width=20mm]{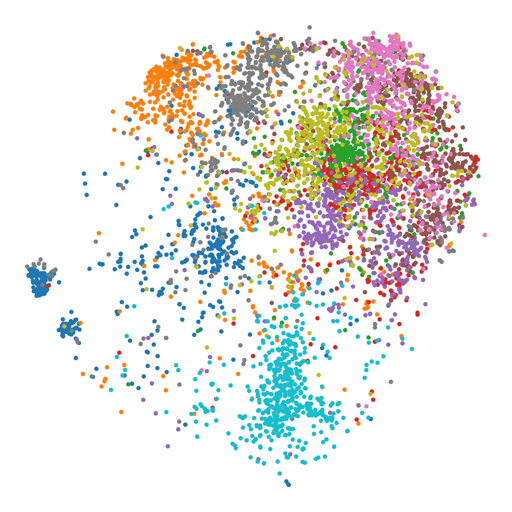} & 
		\includegraphics[width=20mm]{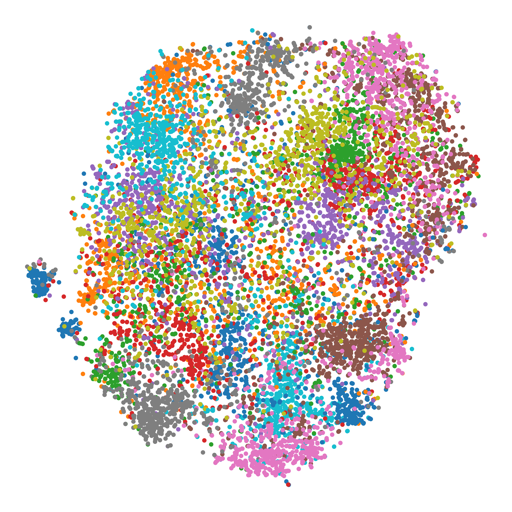} & 
		\includegraphics[width=20mm]{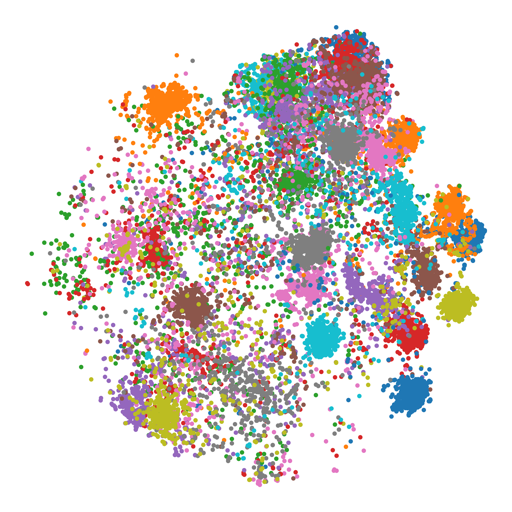} \\
		\includegraphics[width=20mm]{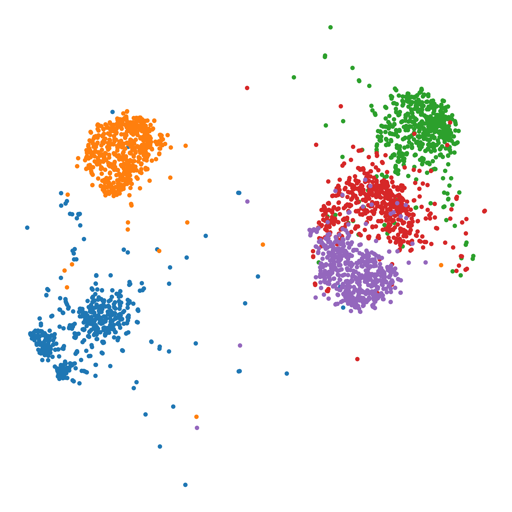} &
		\includegraphics[width=20mm]{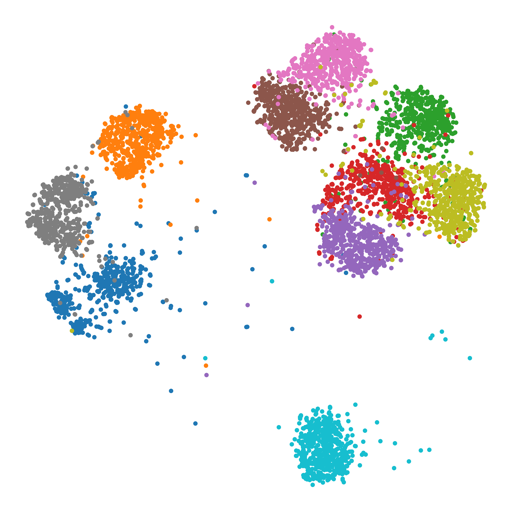} & 
		\includegraphics[width=20mm]{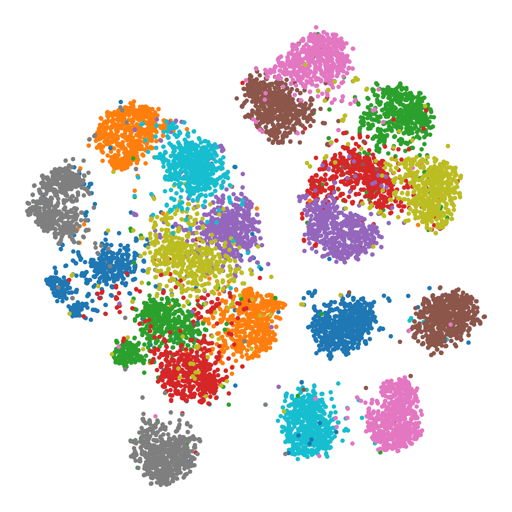} & 
		\includegraphics[width=20mm]{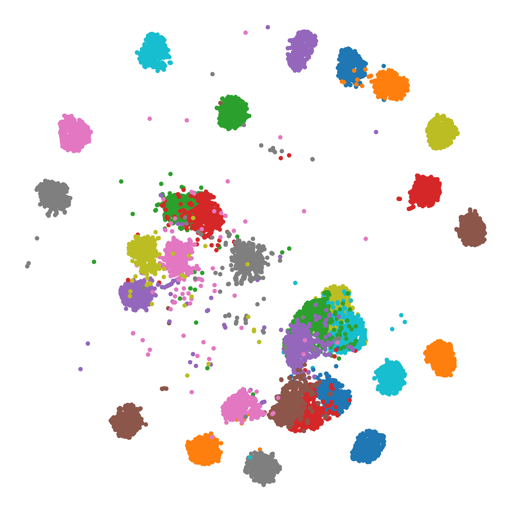} \\
		(a) & (b) & (c) & (d) \\
	\end{tabular}
	\caption{\textbf{TSNE of features}. The top row shows base features after global pooling only whereas the bottom shows features after adaptation on \textit{mini}ImageNet w/ a $5$-way $5$-shot model. \textbf{(a)}, \textbf{(b)} and \textbf{(c)} are embeddings of 5, 10 and 20 classes in the test-set. Finally, \textbf{(d)} are embeddings of 30 random classes in the train-set.}
	\vspace{-3mm}
	\label{fig:feat}
\end{figure}

Here, we evaluate the benefit of co-adapting features in a cross-relational way with the shared attention module $\text{MAB}_h$. \Table{effectmodules} shows the mean accuracy of decomposing our proposed method to assess the effect of omitting or including one of the components. When self-attention is employed on support samples only \ie, $\text{MAB}^{S}_h$, performance drops significantly, especially in the $1$-shot setting where results are below existing state-of-the-art baselines as opposed to a higher $5$-shot accuracy that may imply sensitivity to the number of samples for prototype generation. Interestingly, adaptation of queries ($\text{MAB}^{Q}_h$) alone shows consistent results, yet is marginally lower than the proposed method in the $5$-shot setting. 

Finally, we assess the benefit of using self-attention only similar to FEAT \cite{ye2020few}. However, our implementation applies support-support and query-query ($\text{MAB}^{*}_h$) only without any adaptation, nor do we employ regularization terms in the learning objective. Our findings suggest that self-attention alone does not improve over the baseline models such as ProtoNet (\Table{1}), and also suffers in the extreme $1$-shot setting.  

\subsection{Qualitative Results}

The effects of cross-adaptation of features via the attention scores per branch are shown in \Fig{qualityatt}. We randomly sampled a task on \textit{mini}ImageNet and varied the number of query images in the $5$-way $5$-shot setting \ie, $\{1,2,3\}$ queries with $5$ support images, respectively. For (a), the attention score of a single query will have no effect in adaptation, given its a single sample.  We observed that our method adds more weight to the most similar support samples, and lowered weights for samples with extremely varied background or color. Moreover in (b), due to the presence of multiple objects at a distance (support image w/ $0.08$), less weight was given. Interestingly, the model penalizes the query with a close-up of the \textit{wild dog}. We observed similar trends as query images increased, as in the case (c); querys with \textit{boxes} only were weighted equally, whereas the query with the dog was penalized. Given that our method does not explicity employ visual correspondence to re-weights features, this shows the viability and robustness of our method.   

In \Fig{feat}, we present a visualization of embeddings before and after adaptation. Generally, few-shot models trained episodically can not form accurate clusters of the existing classes due to limited samples in meta-training, unless explicitly enforced. We observed that after applying our strategy on base representations, features were more compact and embeddings were clustered relative to the true class. We owe this to our method adding attention pooled prototypes to the initial features, ensuring either query or support samples are more robust.

\paragraph{Limitations} Our method shows reasonable improvements over recent approaches leveraging instance embeddings only. However, producing interpretable maps that explicitly highlight pixel-level correspondence between images rather than attention scores may be more desirable. Moreover, it may be computationally prohibitive when larger backbones are employed with more heads. Thus, extension of our method to fully-convolutional MAB with spatial features may be interesting, and is left for future research.

\section{Conclusion}
\label{sec:conclusion}

In this work, we revisited discriminative self-attention for few-shot classification via cross-relational adaptation of support and queries. Our work has shown that attending to features in a single view only may be limited, instead cross-adaptation that attends to relevant features via re-weighting with task prototypes can boost performance and generalize better in the cross-domain setting. We also demonstrate state-of-the-art performance without the requirement of transfer learning.

\paragraph{Acknowledgments} This work was supported by the DGIST R\&D program of the Ministry of Science and ICT of KOREA (19-RT-01 and 21-DPIC-08), and IITP grant funded by the Korean government(MSIT) (No.2021-0-02068, Artificial Intelligence Innovation Hub).

%%%%%%%%% REFERENCES
{
    \clearpage
    \small
    \bibliographystyle{ieee_fullname}
    \bibliography{macros,main}
}

% --- supplementary material
%\input{sec/X_supplementary}

% --- uncomment this to read the instructions
%\input{sec/X_instructions}

\end{document}